\documentclass[sigconf]{acmart}

\usepackage{multirow}
\usepackage{subfigure}
\AtBeginDocument{%
  \providecommand\BibTeX{{%
    \normalfont B\kern-0.5em{\scshape i\kern-0.25em b}\kern-0.8em\TeX}}}

\setcopyright{acmcopyright}
\copyrightyear{2021}
\acmYear{2021}
\acmDOI{10.1145/1122445.1122456}

\acmConference[Chengdu '21]{Chengdu '21: ACM Symposium on Neural
  Gaze Detection}{October 20--24, 2021}{Chengdu, Sichuan}
\acmBooktitle{Chengdu '21: ACM Symposium on Neural Gaze Detection,
  October 20-24, 2021, Chengdu, Sichuan}
\acmPrice{15.00}
\acmISBN{978-1-4503-XXXX-X/21/10}

\begin{document}

\title{DRDF: Determining the Importance of Different Multimodal Information with Dual-Router Dynamic Framework}

% \affiliation{%
\author{Haiwen Hong$^{1,2}$,\quad Xuan Jin$^2$,\quad Yin Zhang$^{1*}$,\quad Yunqing Hu$^{1,2}$, \quad Jingfeng Zhang$^{1,2}$, \quad Yuan He$^{2}$, \quad Hui Xue$^{2}$}

\makeatletter
\def\authornotetext#1{
\if@ACM@anonymous\else
    \g@addto@macro\@authornotes{
    \stepcounter{footnote}\footnotetext{#1}}
\fi}
\makeatother
% \authornotetext{Corresponding author.}
\authornotetext{Corresponding author.}

\affiliation{
 \institution{\textsuperscript{\rm 1}College of Computer Science and Technology, Zhejiang University \city{Hangzhou} \country{China}}
 \institution{\textsuperscript{\rm 2}Alibaba Group \city{Hangzhou} \country{China}}
 }
\email{{honghaiwen96, zhangyin98,yunqinghu, zhjf}@zju.edu.cn}
\email{{jinxuan.jx, heyuan.hy, hui.xueh}@alibaba-inc.com}

\def\authors{Yunqing Hu, Xuan Jin, Yin Zhang, Haiwen Hong, Jingfeng Zhang, Yuan He, Hui Xue}

\renewcommand{\shortauthors}{Haiwen Hong, et al.}

\begin{abstract}
In multimodal tasks, we find that the importance of text and image modal information is different for different input cases, and for this motivation, we propose a high-performance and highly general Dual-Router Dynamic Framework (DRDF), consisting of Dual-Router, MWF-Layer, experts and expert fusion unit. The text router and image router in Dual-Router accept text modal information and image modal information, and use MWF-Layer to \textbf{\emph{determine the importance of modal information}}. Based on the result of the determination, MWF-Layer generates \textbf{\emph{fused weights}} for the fusion of experts. Experts are model backbones that match the current task. DRDF has high performance and high generality, and we have tested 12 backbones such as Visual BERT on multimodal dataset Hateful memes, unimodal dataset CIFAR10, CIFAR100, and TinyImagenet. Our DRDF outperforms all the baselines. We also verified the components of DRDF in detail by ablations, compared and discussed the reasons and ideas of DRDF design.
\end{abstract}

% <concept_id>10010147.10010257.10010321.10010333.10010076</concept_id>
% <concept_desc>Computing methodologies~Boosting</concept_desc>
% <concept_id>10002951.10003227.10003251.10003256</concept_id>
% <concept_desc>Information systems~Multimedia content creation</concept_desc>

% \ccsdesc[500]{Computing methodologies~Boosting}
% \ccsdesc[500]{Information systems~Multimedia content creation}



\maketitle
\begin{figure}[]
    \centering
    \includegraphics[width=0.5\textwidth]{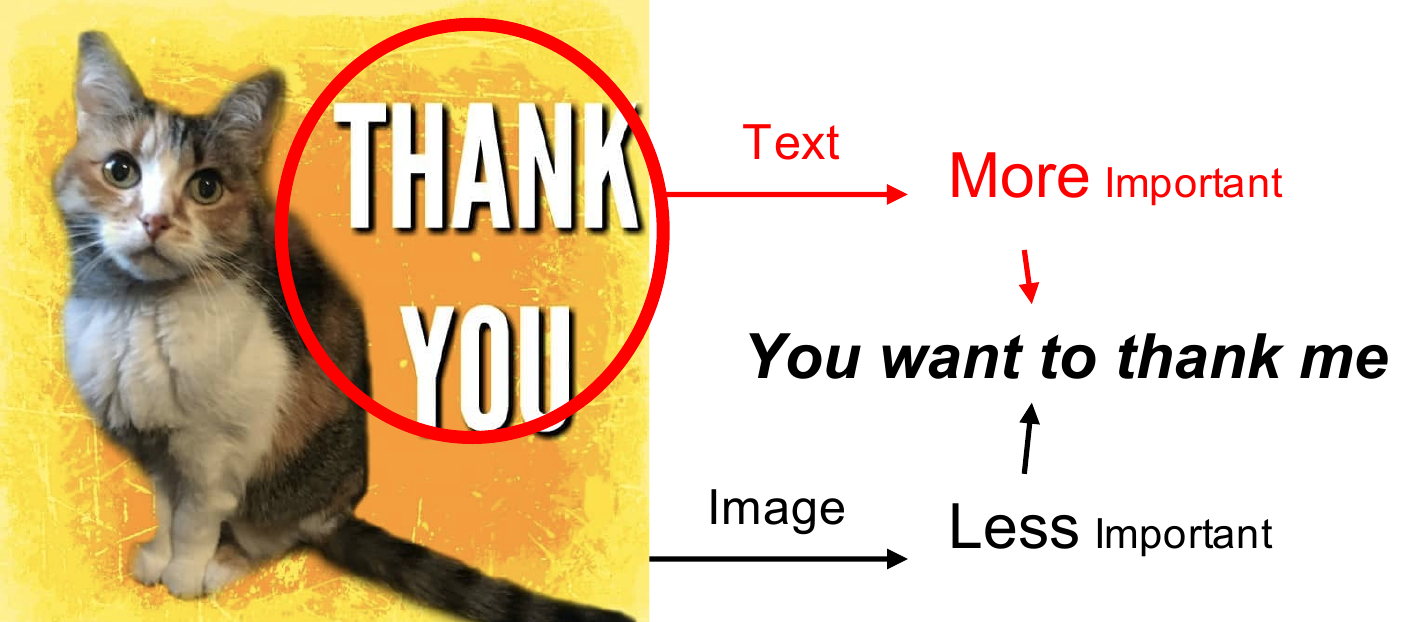}

    \caption{The motivation of DRDF. People can quickly determine that it means ``you want to thank me'', where the more important source of information is the text ``Thank you'', while the cat in the background picture has little impact on people's understanding.}
    \label{fig:motivation}
\end{figure}

\section{Introduction}
Multimodal learning has developed rapidly in recent years, and a number of excellent works have emerged\cite{li2019visualbert,lu2019vilbert,kiela2019supervised}. They extract and combine features from different modalities to obtain richer information. However, most of them have the following drawbacks: (1) ignoring the difference of importance of different modal information, using one fixed architecture to deal with all multimodal problems even though the importance of modal information in different input cases is completely different; (2) researching about single model architecture, and it is difficult to integrate and complement each other's work; (3) poor generality, which cannot be unified for unimodal tasks such as NLP, CV and multimodal tasks.

In our study, we found that the importance of information varies dramatically across modalities even in the same task. In real-world life, humans do not treat information of different modalities equally. For example, when reading a novel with illustrations, the main way we learn the plot of the novel is by reading the text rather than observing the illustrations, while in some comics we tend to pay more attention to the images than to the text. As shown in Figure \ref{fig:motivation}, we can tell at a glance that the meaning of this picture is ``you want to thank me'', while the main information is in the text ``thank you'', and the background of the picture does not give us such clear information, even if we replace the cat in the background image with a dog or a fish, it has almost no effect on the meaning of this meme. Also during the creation of Hateful memes\cite{kiela2020hateful}, it was mentioned that the replacement of text or images in many memes actually does not change its original meaning, which means that in some multimodal samples, the information of one of the modalities will be more important. Even under a specific dataset like Hateful memes, the importance of the modal information in the specific input case inside is very different.

This difference in the importance of modal information is widespread in multimodal tasks, which means that \textbf{\emph{the model is likely to lead to many errors if it treats different modal information equally}}, when facing specifically each different input case even in the same task.

Based on the above motivation, we designed the Dual-Router Dynamic Framework(DRDF) for both multimodal and unimodal learning, which is high-performance, general and modular. It consists of Dual-Router (image router and text router), multiple experts, a MWF-Layer(Modal Weight Fusion Layer) and a expert fusion unit. The image router and text router generate two different sets of weights(called text weights and image weights) based on the different modal information of the input case, and these two weights are fused in the MWF-Layer to form a set of \textbf{\emph{fused weights after the determination of the importance of modal information}}, which will guide the different experts to fuse to get the final result that is most applicable to that input case.

Compared with previous work, our architecture (1) can compare the importance between multiple modalities in MWF-Layer and dynamically fuse to get the results that best fits the current input case; (2) is a highly modular framework that can be perfectly orthogonal to existing work and help existing work further improve performance; (3) is general and can be processed for a large number of models and can be adapted for both multimodal and unimodal tasks.

Experts can be a variety of existing model backbones. The experts have strengths and weaknesses, and Dual-Router and MWF-Layer output weights to guide the complementary integration of the experts, so DRDF can ultimately achieve a better performance than the original individual experts.

Image router and text router in Dual-Router are two simple neural networks that simply look at the input case and make a rough judgment about the modal information. Dual-Router generates two sets of weights, one from the text modality and one from the image modality. In each specific input case, the importance of the two modalities is different, so we design a novel MWF-Layer, which can determine the importance of each modal information based on the data distribution by two types of weights. MWF-Layer will fuse the text weights and image weights according to the importance of the different modal information, so as to produce fused weights that fit the current input case and guide the fusion process of experts to get the best final results in the direction of focusing on the more important modalities.

Experts work like a traditional multimodal model, accepting multiple modal information from the input case and then giving prediction results. So the core of DRDF lies in two data streams of input cases, the first one through Dual-Router, MWF-Layer, and the other one directly through experts and expert fusion unit, to get the prediction result. 

Understandably, Dual-Router and experts can be arbitrarily replaced with any existing available backbone, and when facing unimodal tasks, we can use only one of the router, which makes DRDF modular and general.

In summary, our contributions are as follows:
\begin{itemize}
    \item we propose a novel high-performance and general framework DRDF with a novel Dual-Router adapted to both multimodal and unimodal tasks. We experimentally verified the effectiveness of Dual-Router and explored the impact of continuity on Dual-Router design.
    \item we design a novel MWF-Layer that can determine the importance of different modal information and weight the fusion according to the importance. We verified its effectiveness by experiments.
    \item We extend DRDF on 12 backbones such as VGG16, WideResnet, Resnet, MMBT, Visual Bert, etc., and do experiments on 4 datasets including multimodal datasets and unimodal datasets. DRDF outperforms all the baselines, fully illustrating its high performance and generality.
\end{itemize}

\section{Related work}
\paragraph{Multimodal Works}
In recent years, there has been a lot of works on multimodality. In Visual BERT\cite{li2019visualbert}, unicoder-VL\cite{li2020unicoder}, VL-BERT\cite{su2019vl}, textual and visual information are fused at the beginning, and in ViLBERT\cite{lu2019vilbert}, LXMERT\cite{tan2019lxmert}, the textual and visual information initially go through two separate encoding modules before fusing the different modal information through mutual attention mechanisms. They attempt to transfer some unimodal frameworks such as BERT to the multimodality and to build a generalizable feature learning model in the multimodality. 

In the process of multimodal information fusion, there is quite a bit of work using attention. \cite{hori2017attention} uses attention to fuse the multimodal information and features, and achieves excellent performance in video description task. \cite{ovalle2017gated} proposes GMU based on gated neural network, which can weight the fusion of information from different modalities in the video task.

While the above works are all optimized in the model itself, our work focus on the connection and scheduling between models.
Our work is orthogonal to all the above works, and in our experiments that DRDF can extend these work as backbones, and we use Dual-Router and MWF-Layer to achieve full scheduling of these experts, and then help them achieve a higher performance.

\paragraph{Dynamic Neural Network and Mixture of Experts}
According to this comprehensive and detailed survey\cite{han2021dynamic}, dynamic neural networks and mixture of experts are increasingly popular fields, with excellent works coming out all the time. SENet\cite{hu2018squeeze} slices the traditional network into squeeze and excitation steps, and increases performance with a small extra cost by first descending and then ascending. In DY-CNN\cite{chen2020dynamic}, each layer uses attention to gather multiple input independent convolutional kernels and fuse them, exchanging a small consumption for better performance. \cite{sharma2021long} divides the dataset into manyshot, mediumshot and fewshot according to the number of class inclusions, trains on these subsets respectively to ensure the diversity of the experts, pretrains the baseline on the whole dataset for knowledge transfer, and then different experts for different subsets for finetuning to handle the long-tailed tasks.

Some methods\cite{chen2017sca,woo2018cbam} make the network dynamic by ignoring or cutting certain regions. SKNet\cite{li2019selective} proposes an efficacious module on the basis of channel attention. \cite{zhong2020squeeze} proposes the pixel-group attention to enrich spatial information in SENet. Deformable Kernels\cite{gao2019deformable} resamples the original kernel space to fit the effective receptive field. CondConv\cite{yang2019condconv} uses the routing function to output weights for linear fusion with experts, which can improve performance while keeping the inference costs in an acceptable range. 

Our work is very different from the above work, firstly, the scope of our task is mainly in multimodal tasks, secondly, CondConv only uses the routing function as a weight generator without any meaning, while our Dual-Router is more explainable,  becuase it is designed for text modal and image model respectively, and used MWF-Layer to detect and fuse the importance of modal information, whether the overall design concept or model structure, DRDF is novel. In addition, DRDF is highly modular, experts can be compatible with many existing backbones, without any changes to the backbone internal.

\begin{figure*}[!t]
\centerline{\includegraphics[width=0.85\textwidth]{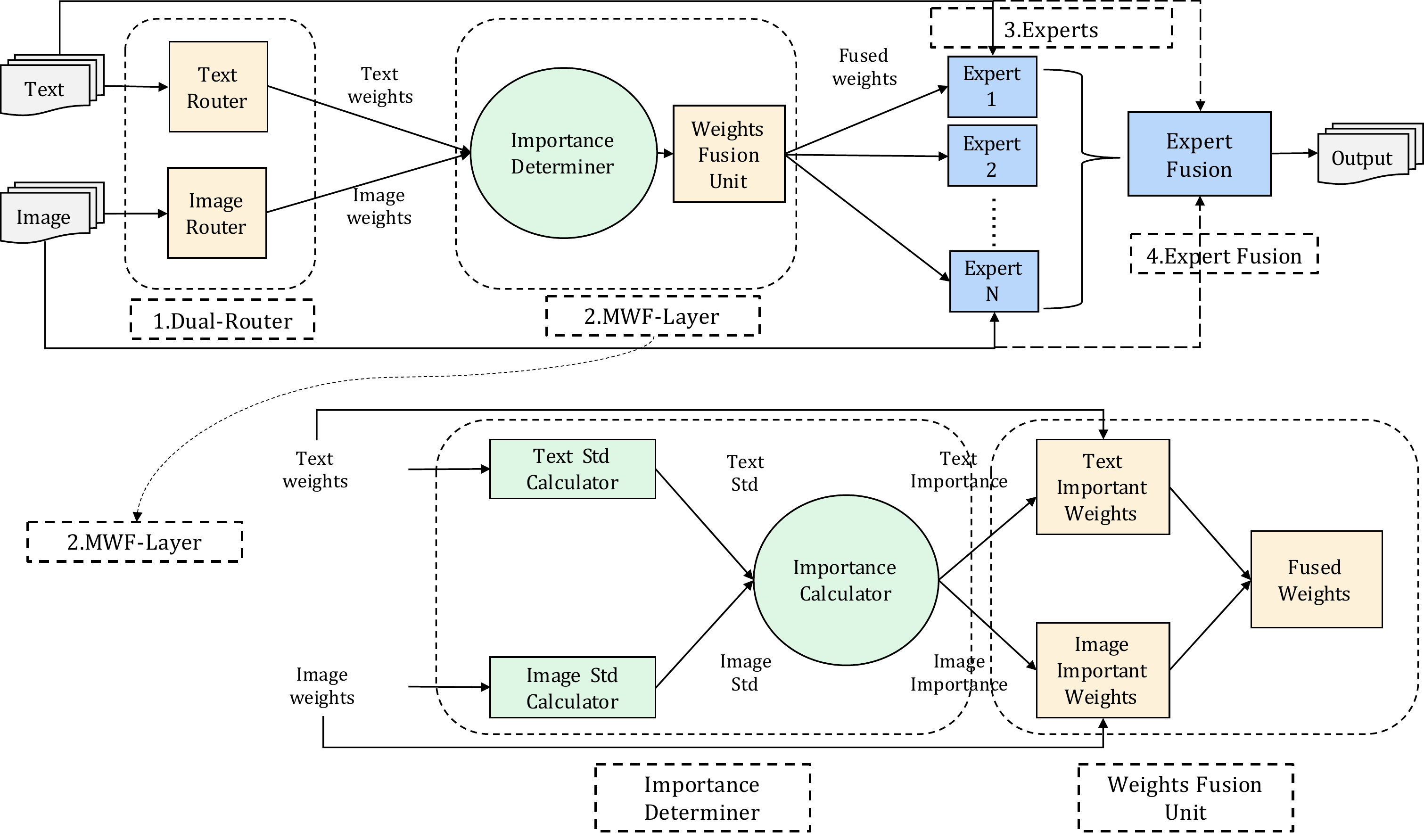}}
\caption{Workflow of the DRDF. The text and image inputs are passed through the text router and image router respectively, generating corresponding text weights and image weights. Two types of weights are determined in the Importance determiner (Text Std Calculator is used to get the standard deviation of weights), and the Weights Fusion Unit fuses them to generate fused weights according to the importance of the modalities. Experts directly processes the input multimodal information to obtain many results. Fused weights instruct experts to perform fusion in expert fusion unit to generate final results. }
\label{fig:overview}
\end{figure*}
\section{Dual-Router Dynamic Framework}
DRDF is a general and high-performance framework, as shown in Figure \ref{fig:overview}, which consists of four parts, and we assume that the number of experts is $N$ for the sake of illustration. (1) Dual-Router contains text router and image router, which accept text and image information of multimodal input, respectively, and generate corresponding text weights and image weights by forward propagation. The dimension here is $N*1$ (without considering the batch size). These weights will be input to the MWF-Layer for importance determination and fusion. The roles of two routers are not for fine classification, but a preliminary analysis of two input modalities, so the router is generally a simple neural network, for example, the text router we choose is LSTM, while the image router we choose is VGG16. (2) The Importance Determiner in MWF-Layer determines the importance of the two modal information by the data distribution of text weights and image weights, and the Weights Fusion Unit fuses them according to the difference in importance to obtain fused weights (with the same dimension of N*1) for experts. We will explain the details of MWF-Layer later.(3) The structure is the same among experts, and the parameter values are different. They are responsible for accepting the input text information and image information, respectively, and outputting the many results. The base performance of the experts determines the overall performance of the whole DRDF.(4) In expert fusion unit, results will perform weighted fusion according to fused weights to get final results. The input cases need to pass through DRDF following two paths of data streams, the first one is used to get the fused weights according to the importance of each modality for the current input case dominated by Dual-Router, MWF-Layer, and experts, and the second one is directly through experts and expert fusion unit.

In addition, when facing a unimodal task such as image classification in CV, the image router can work, while the text router does not accept data, then the fused weights are exactly the same as the image weights, and DRDF can still run normally. Also if we are dealing with an NLP task, then we just need to start the text router and replace the experts with the backbones suitable for the current NLP task so that our framework is also compatible with unimodal NLP tasks. 

When adding a new modality, as long as an existing neural network available for the modal task is found as router and experts (it is not difficult because DRDF can be adapted to almost all the backbones), MWF-Layer can still work and get the fused weights, so the generality of DRDF is excellent.

What can be seen is that the Dual-Router, experts in DRDF can be replaced arbitrarily without any internal changes to the backbones, which makes DRDF work together with many popular models today. The dynamic mechanism of combining Dual-Router and MWF-Layer essentially complements experts' deficiencies so that ultimately achieves performance improvements.

\subsection{Dual-Router}
In DRDF, Dual-Router contains text router and image router, both of which have the same calculation principle as follows:
\begin{equation}
    \omega = \text{Sigmoid}(\text{Router}(x))
\end{equation}
where $x$ is the input, $\omega$ is the text weights or image weights.

In fact, Dual-Router makes a coarse judgment of the multimodal information, and the result of this judgment is output in the form of weights for experts fusion. This means that Dual-Router does not require too fine and accurate judgments.

It is worth mentioning that the direct output results of router cannot be directly used to input into MWF-Layer. Because firstly, the backbone of text router and image router is very different, and the probability distribution of their outputs will also be very different, which is not comparable, and it is not reasonable to let MWF-Layer to make importance comparison directly. Secondly, in the subsequent experts fusion, if the output of router is used directly, there may be negative weights, which is not intuitive in the model fusion. The reason why Sigmoid is used is that DRDF does parameter fusion on the experts, which means that we prefer to get the reasonable weights of each expert instead of deliberately selecting a certain expert with the highest weight, so for Dual-Router, it essentially performs a multi-classification process instead of a single classification. In multi-classification, Sigmoid works better compared to Relu and Softmax. We also verify these in our subsequent experiments.

\subsection{Modal Weight Fusion Layer(MWF-Layer)}
MWF-Layer accepts weights from Dual-Router and judges its importance based on its data distribution, and then generates fused weights based on its importance. During the testing of routers, under our initialization conditions, we find when the router learns poorly, it always tends to go for outputting a set of average parameters, and when it outputs a set of parameters with a particularly pronounced tendency, it tends to achieve a relatively good performance improvement. It is also consistent with human intuition that the parameters output by the router should be distributed enough to help assemble a diverse set of final results. Based on the above motivation, after we tested many computational approaches, we chose the following algorithm:
\begin{equation}
    Importance_{text} =std(\omega_{text}) / (std(\omega_{text}) + std(\omega_{image}))
\end{equation}
\begin{equation}
    Importance_{image} =std(\omega_{image}) / (std(\omega_{text}) + std(\omega_{image}))
\end{equation}
where $std$ is the standard deviation, the above formula holds at $std(\omega_{text}) + std(\omega_{image}) > 0$, and $Importance$ is the importance of the corresponding modal information. If $std(\omega_{text}) + std(\omega_{image}) = 0$, that actually represents $std(\omega_{text}) = std(\omega_{image}) = 0$, which means that text weights and image weights assign exactly the same weight to each expert, so we think the text importance and image importance are consistent at this time, $Importance_{image} = Importance_{text} = 0.5$. Then we need to fuse the weights according to the importance of the modalities, as follows:
\begin{equation}
    \omega_{fused} = \omega_{text} * Importance_{text} + \omega_{image} * Importance_{image}
\end{equation}
$\omega_{fused}$ represents fused weights and will be used for fusion of experts to get final results. While in unimodal tasks, since there is only one modality, it is not necessary to calculate it, and it is directly considered that the $Importance$ of that modality is 1 and the other modality is 0. Then MWF-Layer is also compatible with unimodal tasks.

Unlike ordinary attention-based fusion of modal information and features, the MWF-Layer fusion is based on the weights output by Dual-Router for expert fusion, rather than directly on the modal information itself, which is a mixture of experts fusion. The purpose of its fusion is for subsequent multiple experts to fuse a final results that is more suitable for the current input sample, while the multimodal information fusion in previous work is to allow subsequent models to get more reasonable input vectors, which are fundamentally different.

MWF-Layer is essentially encouraging the diversity of router weights, because in the optimization process, modalities with large variance represent more diverse in their weights, thus acquiring greater importance and more involvement in the training process, which will also be subject to better optimization in the backpropagation process, forming a virtuous circle. We also experimentally validate the effect of MWF-Layer.

\subsection{Experts}
Expert can be a variety of existing backbones, and for specific tasks such as multimodal classification tasks, expert can be a backbone that can handle multimodal classification tasks alone. 

The role of experts is to deal with the input case and give some candidate results for expert fusion unit to fuse to the final results as follows:
\begin{equation}
    o_{n} = E_{n}(x)
\end{equation}
where $E_{n}$ is the $n$-th expert, $o_{n}$ is the output of the $n$-th expert, $x$ is the input case.

It is worth mentioning that the experts can be a single model or a multi-model combination of framework, as long as to ensure that its input and output meet the requirements of the current task. This means that our work is not in direct competition with previous works on multimodal information fusion, and the various multimodal models proposed by previous authors may become experts of our framework, thus obtaining performance gains. 

\subsection{Expert Fusion Unit}
The expert fusion unit weights the fused weights from the previous MWF-Layer, and the expert fusion unit weights the results of each expert output and obtains the final result as follows:
\begin{equation}
    o = \Sigma_{n=0}^{N}o_{n}*\omega_{fused}(n)
\end{equation}
where $o$ is the final output of DRDF, $N$ is the number of experts, $\omega_{fused}$ is the fused weights from MWF-Layer, $o_{n}$ is the result from the $n$-th expert. Dual-Router has actually made a rough classification of the input case, and the experts fusion according to the fused weights is a fine-grained classification for the input cases.

In fact, any fusion approach that is done in a weighted fusion manner can be applied in DRDF. For example, when the expert is a purely linear layer, the parameter fusion instead of result fusion mentioned in CondConv\cite{yang2019condconv} can be applied. In addition, many dynamic network-related fusion methods can be substituted here, which also makes our framework very modular in the concept of the mixture of experts and dynamic neural network.

\section{Experiments}
\subsection{Setup}
We have extended DRDF to a lot of existing models and obtained significant performance improvements on several datasets. We validated both multimodal and unimodal tasks.

In the multimodal task, the dataset we selected is Hateful memes\cite{kiela2020hateful}, a dataset and benchmark centered around detecting hate speech in multimodal memes. Some of the memes in the dataset are original memes from social media, while others are new memes that are similar to the original memes but have very different meanings by manually replacing the background or textual information of the memes. We extended DRDF for backbones such as Late fusion, Concat BERT\cite{kiela2020hateful}, MMBT-Grid, MMBT-Region\cite{kiela2019supervised}, ViLBERT\cite{lu2019vilbert}, Visual BERT\cite{li2019visualbert}, ViLBERT CC\cite{sharma2018conceptual}, Visual BERT COCO. We use AUROC and accuracy as metrics here.
It is worth mentioning that since the testing process of hateful memes needs to be tested online after uploading to the website, for convenience and fairness, we use the metrics on the validation set for all baselines and extended DRDF for comparison.

In the unimodal task, we tested the image classification task, selected datasets are CIFAR10\cite{krizhevsky2009learning}, CIFAR100\cite{krizhevsky2009learning}, TinyImagenet\cite{le2015tiny} (we think the unimodal test is not the core of the paper, but only to support the generality of DRDF, so there is no need to use the time-consuming full Imagenet). We borrowed part of the baselines of NBDT\cite{wan2020nbdt}, and made DRDF extensions on WiderResnet28*10\cite{zagoruyko2016wide}, resnet18\cite{he2016deep}, VGG16\cite{simonyan2015very}. In the image classification task, we only use the image router, and do not need to use the text router.

For the unimodal text classification problem, we perform a DRDF extension to Text BERT\cite{devlin2019bert} and test it on hateful memes. Since Text BERT only accepts text modal information from hateful memes, this can be considered as a DRDF performance test for text modality.

For unimodal tests, we have used accuracy as the metric.

We use the pretrained backbones with noises as experts, and finetune them in downstream datasets. In the above evaluation, we set the text router to be traditional LSTM, the image router to be VGG16, and the number of experts to be 4. 
\begin{table*}[]
\caption{Evaluation results on valid sets on multimodal dataset hateful memes valid set.NN means original model, DRDF means that we use these backbones as experts to get DRDF instances. Text BERT is a unimodal model that accepts only text information in hateful memes, and Late Fusion, Concat BERT, MMBT-Grid, MMBT-Region, ViLBERT, Visual BERT, ViLBERT CC and Visual BERT COCO are all multimodal backbones.}
\resizebox{1\linewidth}{!}{
\begin{tabular}{@{}l|l|rrrrrrrr|l@{}}
\toprule
                                                                       & Model & Late Fusion & Concat BERT & MMBT-Grid & MMBT-Region & ViLBERT & Visual BERT & \begin{tabular}[c]{@{}l@{}}ViLBERT \\ CC\end{tabular} & \begin{tabular}[c]{@{}l@{}}Visual BERT \\ COCO\end{tabular} & Text BERT \\ \midrule
\multirow{2}{*}{NN}                                                    & AUROC &     65.97        &      65.25       &      68.57     &    71.03         &   71.13      &     70.60        &          70.07                                             &           73.97                                                  &    64.65       \\
                                                                       & Acc   &     61.53        &    58.60         &    58.20       &      58.73       &     62.20    &    62.10         &          61.40                                             &        65.06                                                     &     58.26      \\ \midrule
\multirow{2}{*}{\begin{tabular}[c]{@{}l@{}}DRDF\\ (Ours)\end{tabular}} & AUROC &       66.98      &     67.12        &     70.24      &   72.13          &   72.34      &      71.10       &      71.09                                                 &       74.80                                                      &     66.88      \\
                                                                       & Acc   &     64.82        &      59.81       &     61.67      &  60.74           &   64.26      &      62.59       &    62.22                                                   &                65.19                                             &       60.93    \\ \midrule
\multirow{2}{*}{\begin{tabular}[c]{@{}l@{}}Improvements\end{tabular}} & AUROC &       1.01      &     1.87        &     1.67      &   1.10          &   1.21      &      0.50       &      1.02                                                 &       0.83                                                      &     2.23      \\
                                                                       & Acc   &     3.29        &      1.21       &     3.47      &  2.01           &   2.06      &      0.49       &    0.82                                                   &                0.13                                             &       2.67    \\ \bottomrule
\end{tabular}}
\label{tab:multi results}
\end{table*}

\begin{figure}[!t]
    \centering
    \subfigure[AUROC]{
    \includegraphics[width=0.2\textwidth]{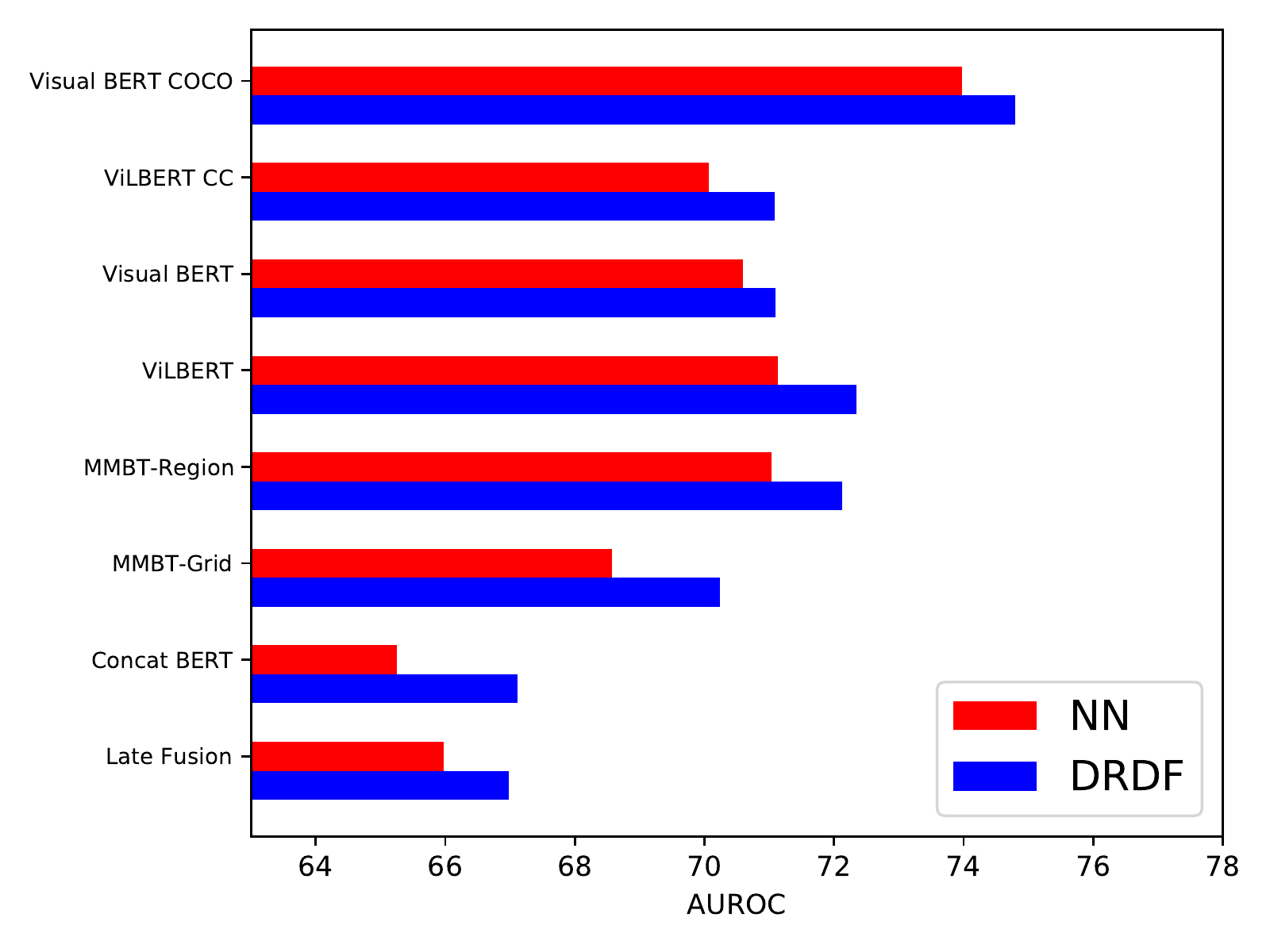}
    \label{subfig:auroc}
    }
    \hspace{0.1in}
    \subfigure[Accuracy]{
    \includegraphics[width=0.2\textwidth]{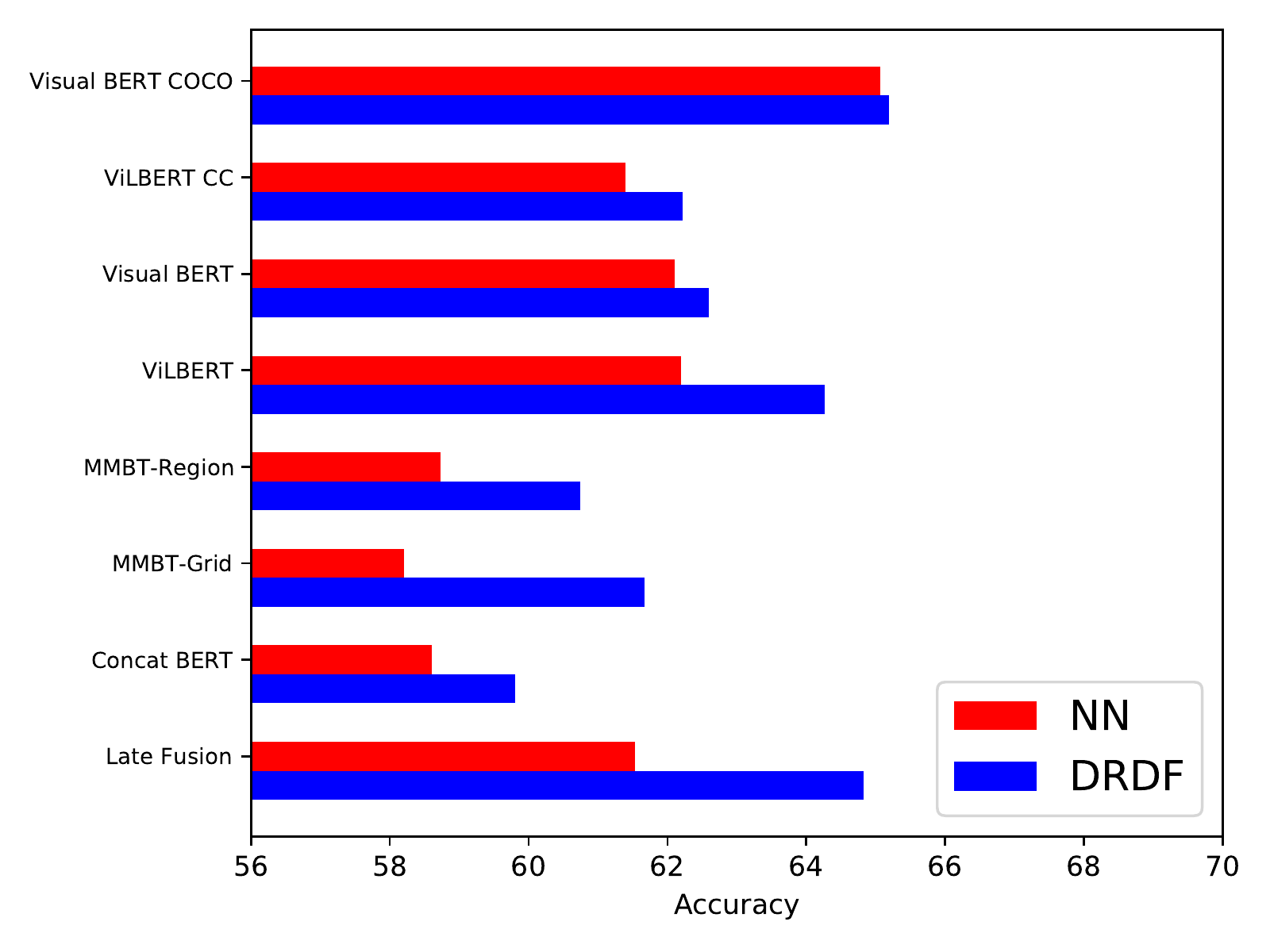}
    \label{subfig:acc}
    }
    \caption{The AUROC and accuracy results chart of multimodal backbones on hateful memes valid set. NN means original model, DRDF means that we use these backbones as experts to get DRDF instances. Text BERT is a unimodal model that accepts only text information in hateful memes, and Late Fusion, Concat BERT, MMBT-Grid, MMBT-Region, ViLBERT, Visual BERT, ViLBERT CC and Visual BERT COCO are all multimodal backbones.}
    \label{fig:multimodal}
\end{figure}
\subsection{Multimodal Results}
Table \ref{tab:multi results} and Figure \ref{fig:multimodal} shows the results. It can be seen that after our DRDF extension, the AUROC and accuracy results of all the backbones have been significantly improved.

In terms of AUROC, the improvements of Concat BERT and MMBT-Grid are the highest, 1.87 and 1.67 respectively, the common point of both is that both use BERT for text feature extraction while accepting the original images directly instead of the extracted image features for image modality. The improvement of MMBT-Region is smaller than that of MMBT-Grid. The model structure of both is almost identical, and the difference is that MMBT-Grid receives the original image as input directly, while MMBT-Region receives the extracted image features as input. The image router also receives the original images instead of the extracted features, so its output weights are more suitable for models that use the original images as the input of the image modality, which is also an intuitive result.

Besides, we found that Visual BERT and Visual BERT COCO have the lowest AUROC improvements of 0.50 and 0.83. It is worth mentioning that both backbones, especially Visual BERT COCO, achieved the best result of 73.97 on hateful memes, which means that the model performs more comprehensively than others. DRDF essentially uses experts fusion to make up for the shortcomings of the experts, and for the relatively comprehensive backbone, the degree of improvement is smaller because there are fewer places to make up. 

In terms of accuracy, we can see more significant improvements. Late fusion and MMBT-Grid have the highest improvements, reaching 3.29\% and 3.47\%, and MMBT-Grid still has a larger improvement than MMBT-Region, which also confirms the difference between the original image or the image features of the modal information mentioned above. Late fusion has a simple structure and its backbone achieves a performance that is downstream in the baseline, which means that it may have many defects that can be filled in as an expert, thus making the role of DRDF very obvious. The accuracy improvements of Visual BERT and Visual BERT COCO are still insignificant at 0.49\% and 0.13\%, which also indicates that their good performance on the hateful memes dataset makes the DRDF role insignificant. VilBERT and VilBERT CC as similar backbone but the difference in accuracy improvements is larger, 2.06\% and 0.82\% respectively, which may be the effect of the pretrained model of VilBERT CC, pretraining brings more comprehensive performance, making the improvements of DRDF are not obvious enough compared to the original backbone.
Overall, the effect of DRDF is impressive, and we have verified through these extensive backbone that DRDF can work in a large number of existing models, and they do not form a direct competition but an orthogonal complementary relationship, our DRDF can easily help other models and frameworks to achieve a better performance.

\subsection{Unimodal Results}
\begin{table}[]
\caption{The accuracy results of image classification tasks on CIFAR10,CIFAR100 and TinyImagenet. The metric is accuracy on the test set. NN means original model, and DRDF means that we use these backbones(WiderResnet28*10, Resnet18 and VGG16) as the experts. We only use image router VGG16 here instead of Dual-Router, and the number of experts is 4. }
\resizebox{1\linewidth}{!}{
\begin{tabular}{@{}llrrr@{}}
\toprule
                                                                                            & Datasets                          & WiderResnet28*10 & Resnet18 & VGG16 \\ \midrule
\multicolumn{1}{l|}{\multirow{3}{*}{NN}}                                                    & \multicolumn{1}{l|}{CIFAR10}      &         97.62         &      94.97    &  93.46     \\
\multicolumn{1}{l|}{}                                                                       & \multicolumn{1}{l|}{CIFAR100}    &          82.09         &     75.92     &  70.39     \\
\multicolumn{1}{l|}{}                                                                       & \multicolumn{1}{l|}{TinyImagenet} &  67.65                &     64.13     &  53.28     \\ \midrule
\multicolumn{1}{l|}{\multirow{3}{*}{\begin{tabular}[c]{@{}l@{}}DRDF\\ (Ours)\end{tabular}}} & \multicolumn{1}{l|}{CIFAR10}      &     97.80             &    95.42      &   94.21    \\
\multicolumn{1}{l|}{}                                                                       & \multicolumn{1}{l|}{CIFAR100}     &     83.52             &    76.95      &  71.52     \\
\multicolumn{1}{l|}{}                                                                       & \multicolumn{1}{l|}{TinyImagenet} &   69.47               &     67.12     &  56.14     \\ \bottomrule
\end{tabular}}
\label{tab:uni results}
\end{table}
We have tested the adaptability of DRDF to unimodal, where the text modal test is shown in the column "text BERT" in Table \ref{tab:multi results}, we can see that the improvement of AUROC reaches 2.23, and the improvement of accuracy reaches 2.67\%. .

In addition, Table \ref{tab:uni results} shows the effect of DRDF on CV task, image classification, the evaluation metric is accuracy, we can see that DRDF shows considerable improvement on all backbone and all datasets.

With the backbone WideResnet28*10, our DRDF achieves 97.80\%, 83.52\%, 69.47\% on three datasets. On CIFAR10, DRDF outperforms WideResnet28*10 by 0.18\%, on CIFAR100, DRDF achieves accuracy 1.43\% higher than WideResnet28*10 and on TinyImageNet the DRDF outperforms the WideResnet28*10 by 1.82\%.

With Resnet18, our DRDF achieves 95.42\%, 76.95\%, 67.12\%. DRDF outperforms ResNet18 by 0.45\%, 1.03\%, 2.99\% on three datasets. 

With VGG16, our DRDF achieves 94.21\%, 71.52\%, 56.14\%, outperforms original VGG16 by 0.75\%, 1.13\%, 2.86\% on three datasets.

The above experiments fully illustrate the generality of DRDF for unimodal and multimodal tasks. This shows that the text router and image router in Dual-Router are also feasible to use alone, while in unimodal, DRDF exists as a mere MoE framework, which improves the performance by providing fused final results for each specific input sample.

\subsection{Ablations}
\subsubsection{Dual-Router ablations}
\begin{table}[]
\caption{Ablations of Dual-Router with different activation functions on hateful memes valid set with MMBT-Grid as experts backbone. The number of experts is 4, text router is LSTM and image router here is VGG16. NN means the original backbone.}
\resizebox{1\linewidth}{!}{
\begin{tabular}{@{}l|lll|lll@{}}
\toprule
             & \multicolumn{3}{c|}{Acc} & \multicolumn{3}{c}{AUROC} \\ \midrule
             & Sigmoid & Softmax & Relu & Sigmoid  & Softmax & Relu \\ \midrule
NN           & \multicolumn{3}{c|}{58.20}    & \multicolumn{3}{c}{68.57}      \\
Text Router  &     59.26    &    61.30     &  59.81    &       68.90   &     67.11    &    67.27  \\
Image Router &     60.93    &    55.93     &  59.44    &   70.12       &     67.49    &    64.09  \\
Dual Router  &     61.67    &   58.33      &  60.19    &     70.24     &    68.24     &    66.97  \\ \bottomrule
\end{tabular}}
\label{tab:router}
\end{table}
\begin{figure}[!t]
    \centering
    \subfigure[AUROC]{
    \includegraphics[width=0.2\textwidth]{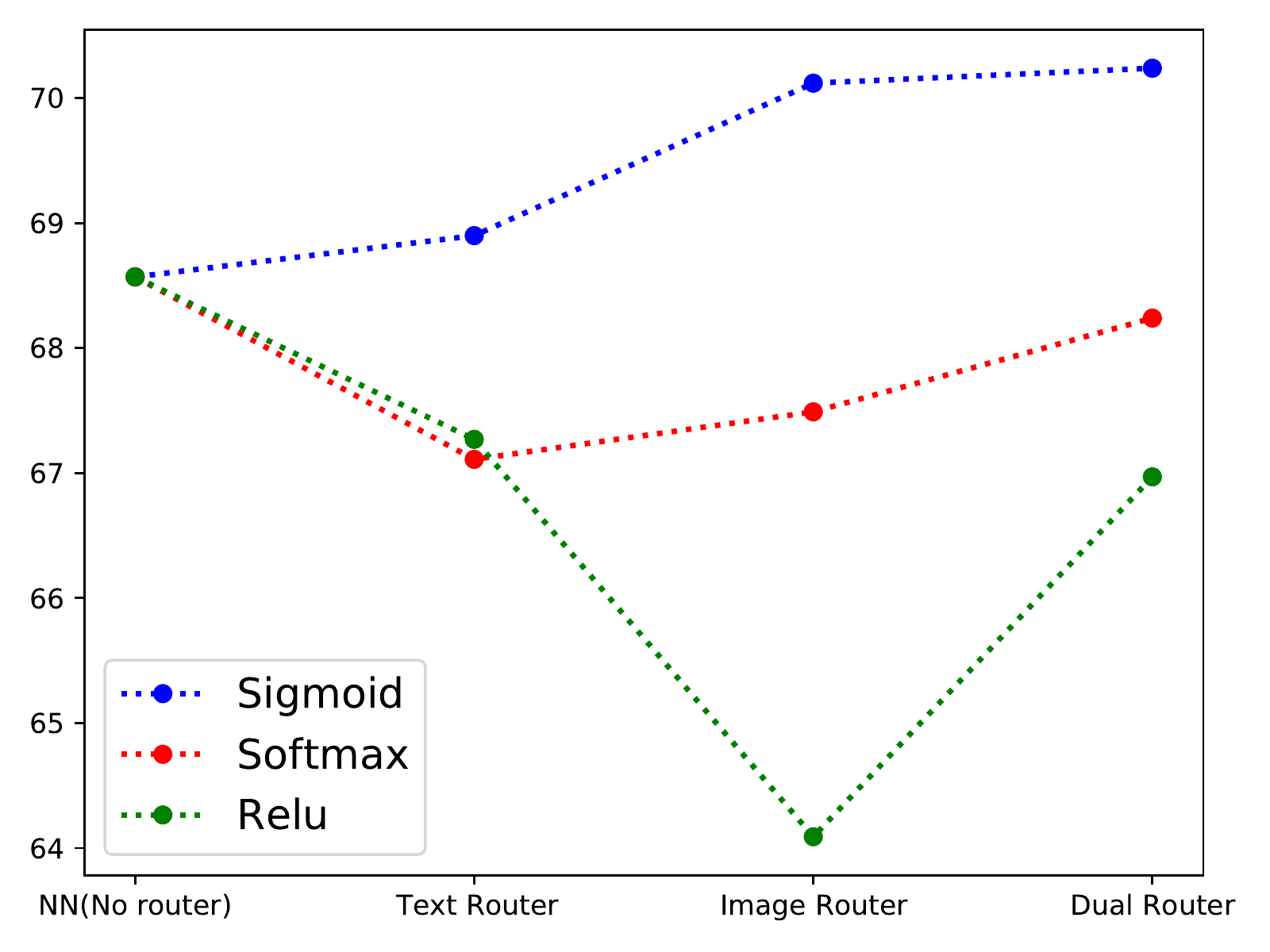}
    \label{subfig:acc_act}
    }
    \hspace{0.1in}
    \subfigure[Accuracy]{
    \includegraphics[width=0.2\textwidth]{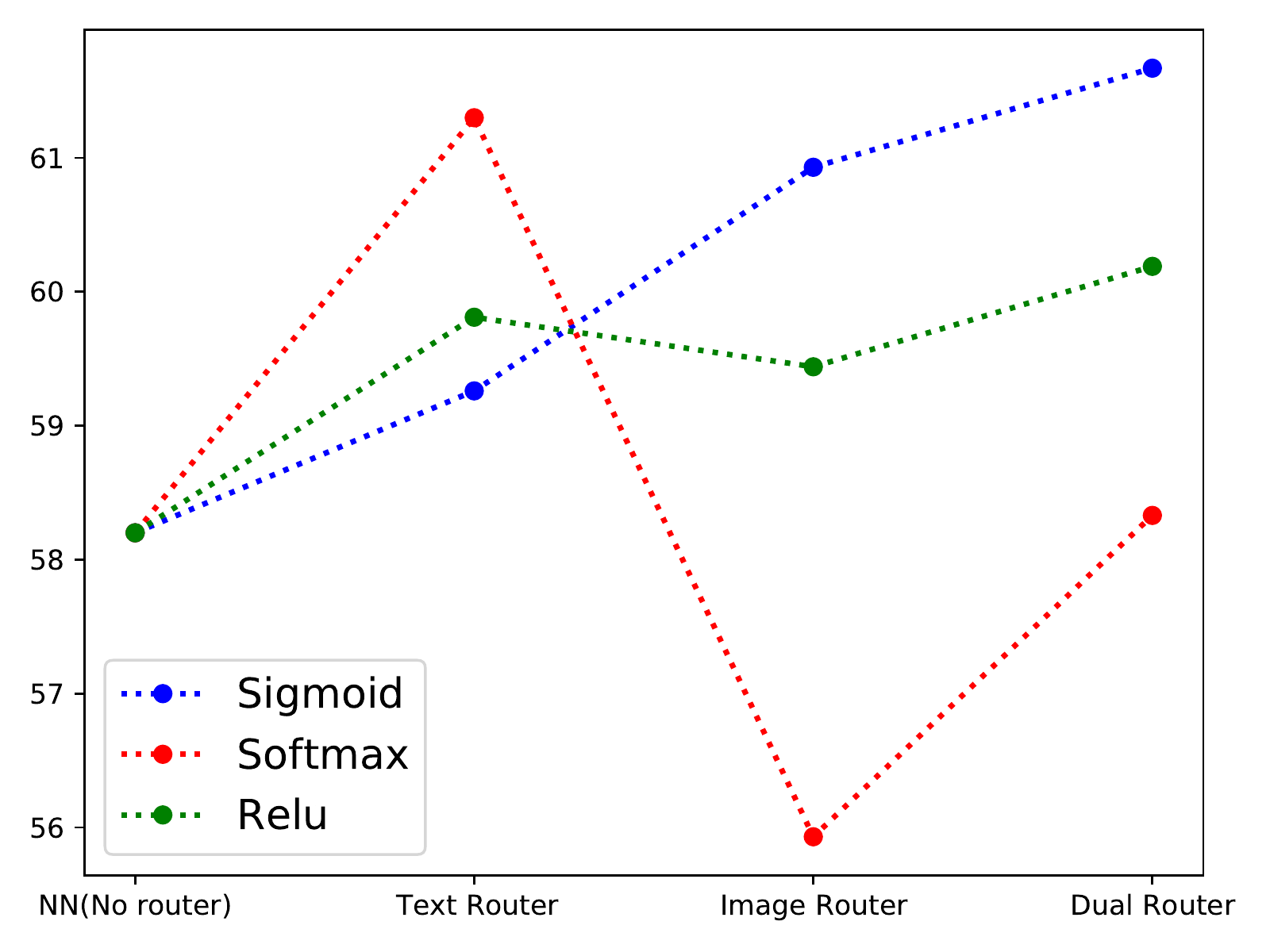}
    \label{subfig:auroc_act}
    }
    \caption{The trend chart of ablations of routers with different activation functions on hateful memes with MMBT-Grid as experts backbones. The left sub-figure and right sub-figure both shows that dual router with Sigmoid works best. }
    \label{fig:router_act}
\end{figure}

To verify the effects of Dual-Router, we do ablations on hateful memes with MMBT-Grid as experts backbone. The number of experts is 4, text router is LSTM and image router here is VGG16. We compared, for different activation functions, the results of using only the text router, only the image router and the Dual-Router. Table \ref{tab:router} and Figure \ref{fig:router_act} shows the results.

For activation function, Sigmoid is the best, probably because Softmax is suitable for a single choice, there is a certain degree of mutual exclusivity among the weights, which will widen the gap between the output parameters, and the output of router needs to allow multiple high weights to coexist, so as to combine a more diverse and accurate mixed model. And this is the advantage of Sigmoid. Similarly, ReLu's direct zeroing of negative parameters will result in serious loss of weights information.

From the router perspective, the introduction of any router under the Sigmoid activation function can give a performance boost, which validates one of the ideas of DRDF, that is, providing a suitable fused final result for each specific input case is able to indeed make up for some of the original shortcomings of backbone and ultimately improve performance. Dual Router outperforms text router and image router by 2.41\% and 0.74\% in accuracy, 1.34 and 0.12 in AUROC, indicating that Dual- Router's design is suitable for multimodal tasks. In addition, we can see that the overall performance of image router is better than that of text router, probably because the model complexity of VGG16 is higher than that of LSTM. But the fusion of the two into Dual-Router can get better results, which shows that Dual-Router is not simply a stack of modal information, it can consider the information of the two modalities to output more reasonable weights, which is the core idea of the DRDF design.

\subsubsection{Discrete router or continuous router}
\begin{table}[]
\caption{Three different router attempts, single-choice gate means that only one expert is selected at a time, multi-choice gate means that more than one expert can be selected for fusion at a time, but the output weight is only the difference between 0 and 1. It's on hateful memes with MMBT-Grid as experts backbones. Text router is LSTM and image router is VGG16. NN means the original backbone.}
\resizebox{1\linewidth}{!}{
\begin{tabular}{@{}l|lll@{}}
\toprule
             & \multicolumn{3}{c}{AUROC}                            \\ \midrule
             & Single-choice Gate & Multi-choice Gate & Continuous(Ours) \\ \midrule
NN           & \multicolumn{3}{c}{68.57}                                 \\
Text Router  &        66.23         &         67.11          &      68.90          \\
Image Router &        67.87         &         67.14          &        70.12        \\
Dual Router  &      68.45           &         67.22          &       70.24         \\ \bottomrule
\end{tabular}}
\label{tab:discrete}
\end{table}

\begin{figure}[!t]
    \centering
    \subfigure[Different gate]{
    \includegraphics[width=0.2\textwidth]{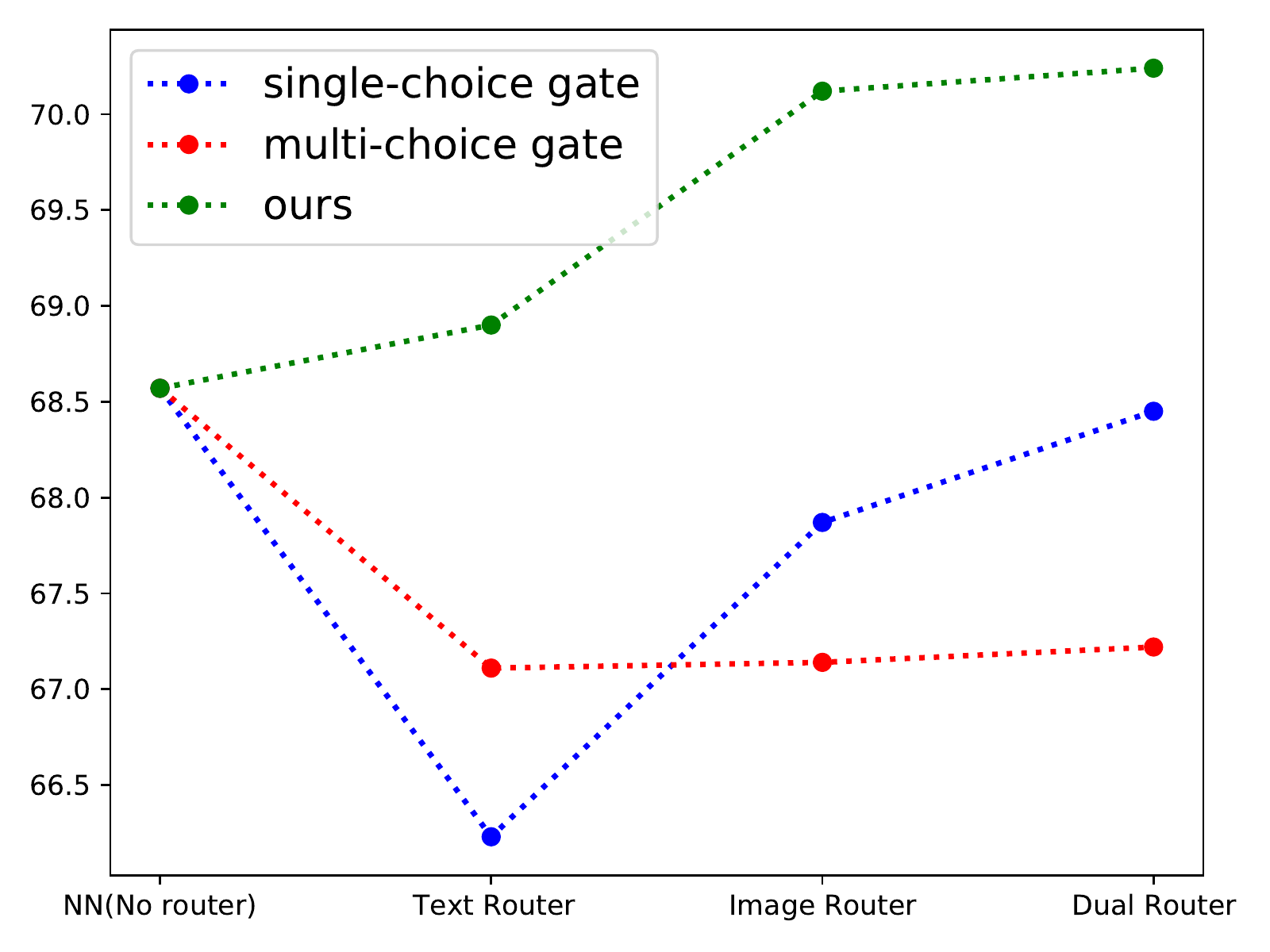}
    \label{subfig:gate}
    }
    \hspace{0.1in}
    \subfigure[Different router]{
    \includegraphics[width=0.2\textwidth]{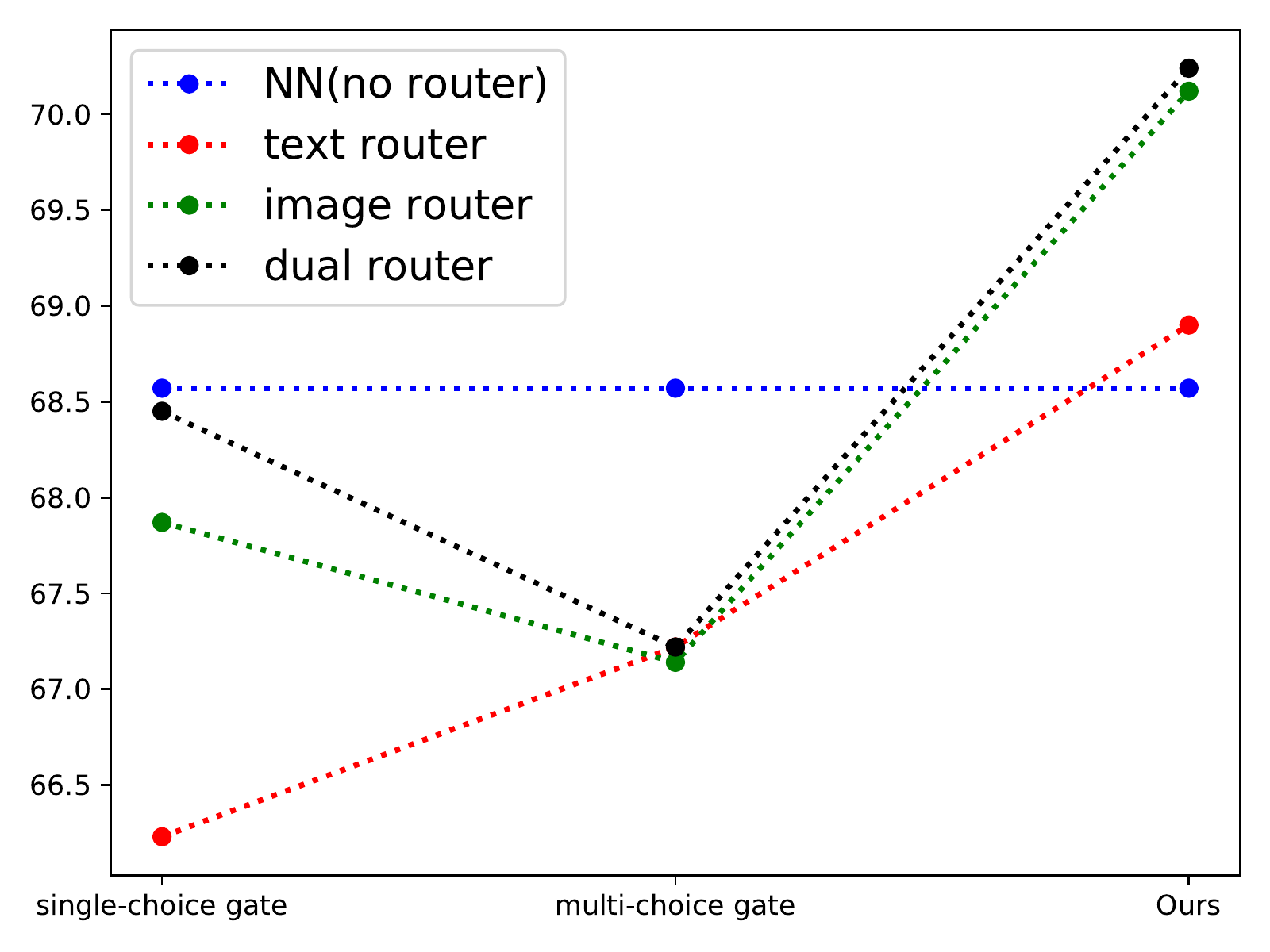}
    \label{subfig:routers}
    }
    \caption{The trend chart of ablations of three different router attempts. The left sub-figure shows that our continuous gate fusion policy works best, while the right sub-figure shows that the dual router works best. }
    \label{fig:discrete}
\end{figure}

The Dual-Router used by DRDF outputs continuous weights and thus fuses the results of the experts based on the weights. We also do tests with discrete routers to verify the effectiveness of continuous routers. 

As shown in the Table \ref{tab:discrete} and Figure \ref{fig:discrete}, we design three different routers, where the single-choice gate will take the largest weight on the fused weights output from the MWF-Layer and set it to 1 (or randomly take the maximum of the tied values if they are tied), and set the rest of the weights to 0, i.e., only one expert is activated at a time. The multi-choice gate will take all the values greater than 0.25 and set them to 1 on the fused weights output by MWF-Layer, and set the rest of the weights to 0 (if there is no value greater than 0.25 then take the maximum bit and set it to 1), i.e. multiple experts can be activated and fused at a time.

We find that our continuous router results outperform the other two by 1.79 and 3.02, respectively. When text router and image router are started simultaneously, single-choice gate performs better than multi-choice gate, and we see that the Dual-Router result for single-choice gate is very close to the original backbone results.

The reason may be that the single-choice gate may fall into a vicious circle in the middle of the training period, i.e., after selecting a certain expert many times, the expert is well trained, which causes the single-choice gate to be more inclined to select it, thus the final training result is close to the result of training a backbone alone. Although multi-choice can be trained by activating multiple experts at the same time, it also leads to the fact that each expert is not sufficiently trained and the final result is inferior to that of single-choice gate.

And obviously continuous-router is the right choice, probably because, firstly, the neural network with continuous values will be easier in training, while Dual-Router is close to the input and far from the output, which is inherently harder to optimize, and will be more difficult to train if discrete-router design is adopted. Secondly, the core idea of DRDF is to combine the unique fused weights for each specific multimodal input case according to the importance of its individual modalities, and only continuous weights can combine infinite possibilities to meet our motivation. Finally, discrete routers are demanding for the ability of the expert, because they cannot combine effective networks, and only rely more heavily on the experts. To the extreme, the single-choice gate is completely dependent on the ability of the expert, and does not help the experts to achieve better results.

\subsubsection{MWF-Layer ablations}
\begin{table}[]
\caption{Ablations of MWF-Layer on hateful memes with MMBT-Grid as experts backbone. The number of experts is 4, text router is LSTM and image router is VGG16. NN means the original backbone. Average represents a direct summation of the weights, and Multiply represents a direct multiplication of the weights.}
\begin{tabular}{@{}l|ll@{}}
\toprule
          & Acc & AUROC \\ \midrule
NN          &  58.20   &   68.57    \\
Average   &   60.37  &    69.12   \\
Multiply  &   57.22  &    65.64   \\
MWF-Layer &  61.67   &   70.24    \\ \bottomrule
\end{tabular}
\label{tab:mwf}
\end{table}
To verify the effects of MWF-Layer, we do ablations on hateful memes with MMBT-Grid as experts backbone. The number of experts is 4, text router is LSTM and image router here is VGG16. 
We compared various approaches to form fused weights, and the results are shown in the Table \ref{tab:mwf}, where Average represents the direct summation of text weights and image router, and Multiply represents the direct multiplication.

What can be seen is that the effect of Average is better than the original backbone and Multiply is poorer. The point of improvement here may be because Dual-Router can make a rough judgment on the modal information, and text weights and image weights have a certain guiding effect on the fusion of experts. When these approaches cannot determine which modal information is more important, Average gives the same importance to both modal information and does not change the weights much, so the final effect of Dual-Router itself makes the performance improved. And Multiply brings more changes, so the final effect is poor.

This fully illustrates the effect of MWF-Layer, which can output fused weights by deciding the importance of information of different modalities, and finally guide the experts to fuse to get the final results that is more suitable for the current input sample.

\subsubsection{Experts ablations}

\begin{table}[]
\caption{Ablations of the experts on hateful memes with MMBT-Grid as experts backbones. NofE means the number of experts. NN means the original backbone. }
\begin{tabular}{@{}l|ll@{}}
\toprule
NofE   & Acc & AUROC \\ \midrule
NN &  58.20   &  68.57     \\
2  &  60.56   &    67.19   \\
3  &   63.52  &    69.39   \\
4  &   61.67  &    70.24   \\
8  &   62.22  &    69.78   \\ \bottomrule
\end{tabular}
\label{tab:experts}
\end{table}
\begin{figure}[!t]
    \centering
    \subfigure[AUROC]{
    \includegraphics[width=0.2\textwidth]{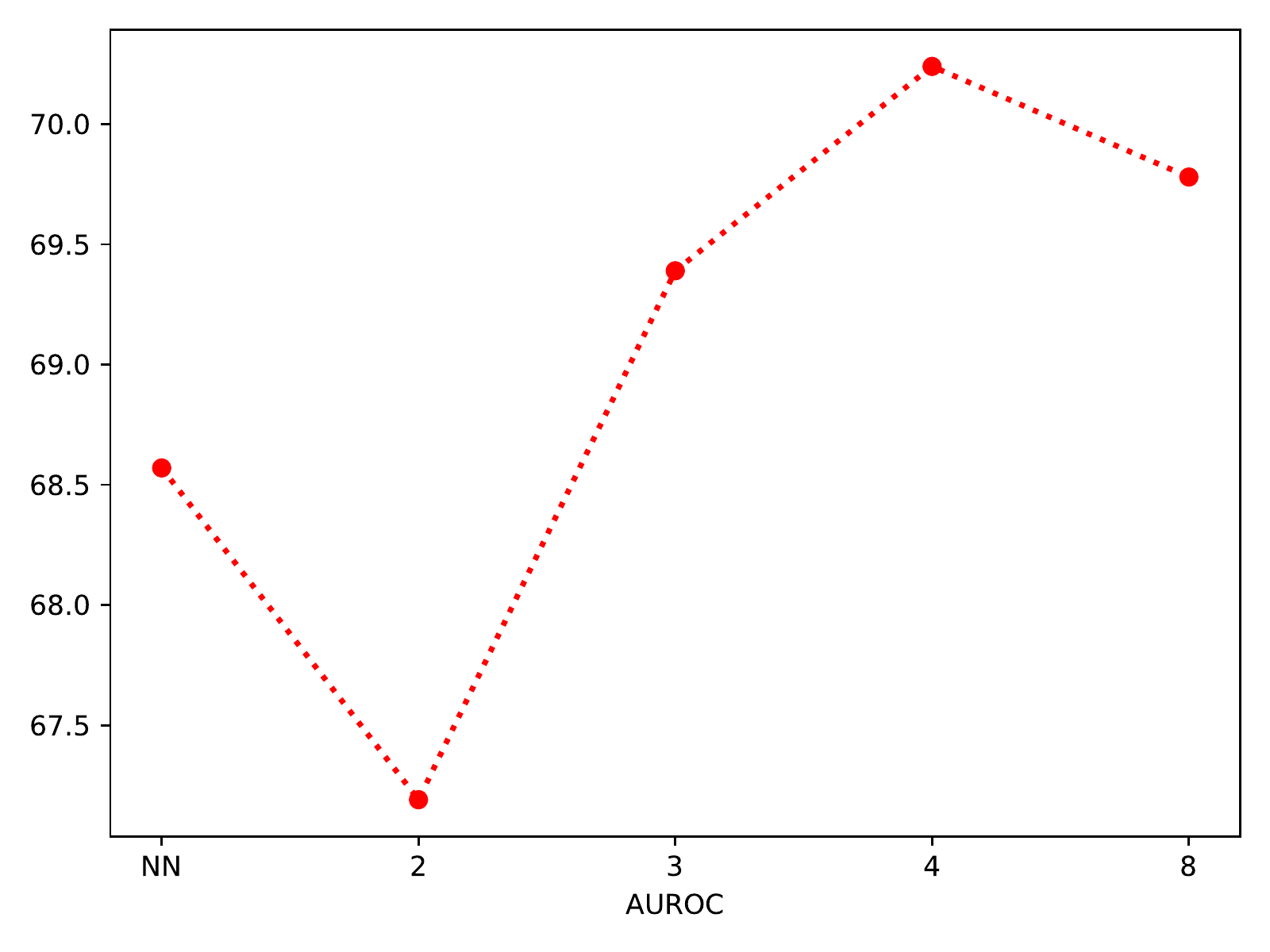}
    \label{subfig:auroc}
    }
    \hspace{0.1in}
    \subfigure[Accuracy]{
    \includegraphics[width=0.2\textwidth]{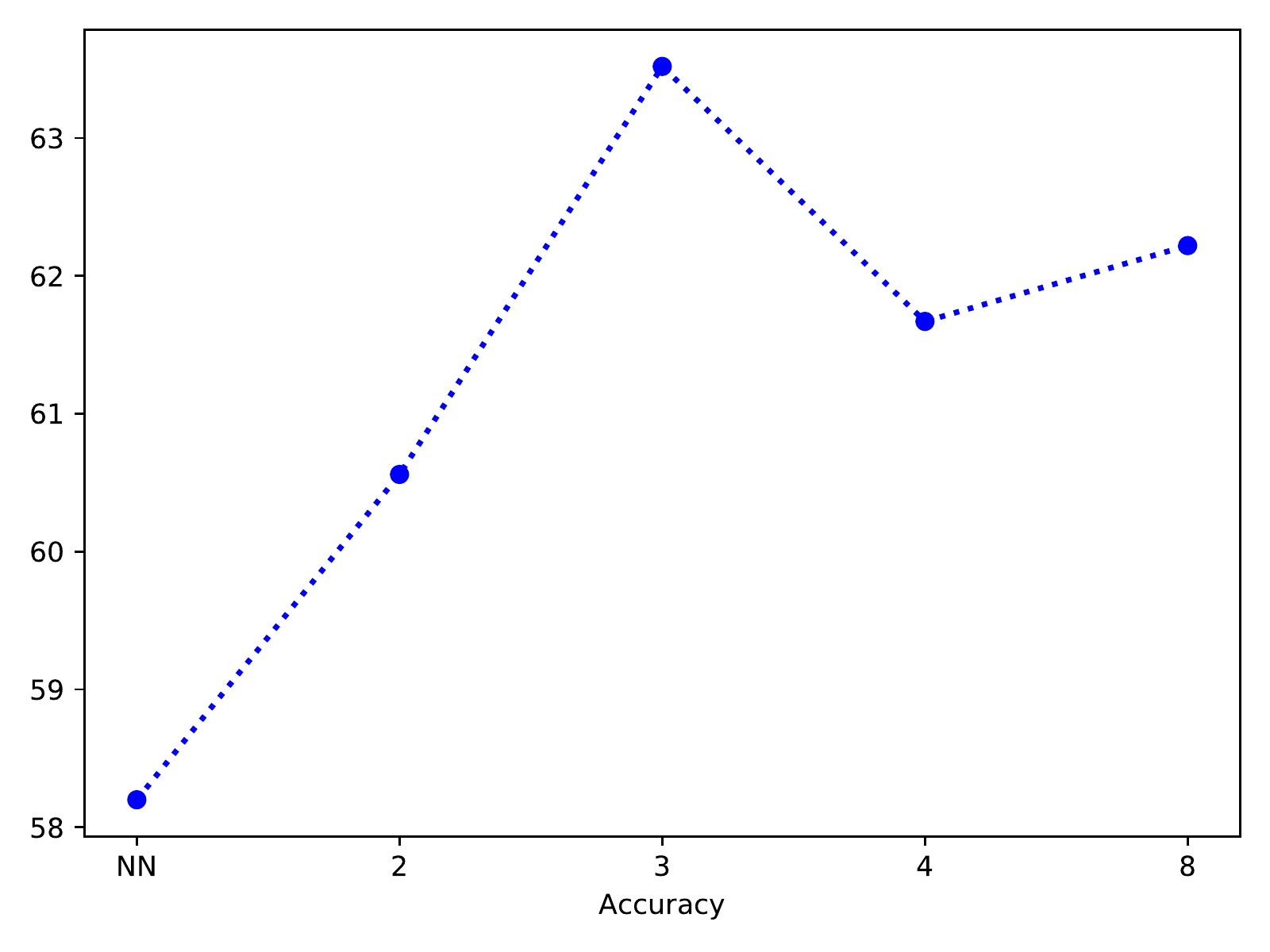}
    \label{subfig:acc}
    }
    \caption{The trend chart of ablations of the experts on hateful memes with MMBT-Grid as experts backbones. NN means the original backbone.}
    \label{fig:noe}
\end{figure}
To verify the effects of experts, we do ablations on hateful memes with MMBT-Grid as experts backbone. Text router is LSTM and image router here is VGG16. 
We compared the DRDF with different number of experts, and the results are shown in the Table \ref{tab:experts} and Figure \ref{fig:noe}.

When the number of experts is 2, accuracy is improved by 2.36\% compared to backbone, while AUROC is reduced by 1.38. The improvement effect of DRDF is not obvious enough, probably because two experts are more difficult to train than one, and the performance of each individual expert decreases. We have verified this point, and found that in DRDF, the performance of almost every single expert is lower than the performance of the original backbone. The improvement of the overall number of parameters generated by two experts is not obvious enough compared to multiple experts, and the improvement of the representation power is not obvious enough, so there is a decline in performance at the stage of experts generating fused final results. This performance degradation is more obvious than the improvement brought by the dynamic mechanism of Dual-Router and MWF-Layer, so eventually, when the number of experts is small, the DRDF effect may be poor. 

After the number of experts increases, we can see that the DRDF effect is improved obviously. Although the effect of individual experts decreases, because the number of experts is larger, its overall number of parameters is larger, and the representation power is stronger, finally reaching the performance increase.

But the number of experts is not as large as it is, we can see that the accuracy of DRDF reaches the highest 63.52\% when the number is 3, which is 5.32\% higher than the original backbone and 1.30\% higher than the number is 8, while the AUROC of DRDF reaches the highest 70.24 when the number is 4, which is 1.67 higher than the original backbone, and 0.46 higher than the number of 8. 

This may be because, in the process of inference, the increase in experts is essentially an increase in the number of parameters that can be selected in the entire DRDF dynamic network. As with traditional networks, the number of DRDF parameters is too large and the dataset is too small, then overfitting is likely to occur, resulting in a number of experts that cannot be increased indefinitely.

Since AUROC is the main metric on the hateful memes dataset, and we do not need an excessive number of experiments in terms of training time cost, the number of experiments we use in our experiments is 4.

\section{Conclusion}
We believe that the model should have the ability to judge the importance of different modal information for different input cases in a multimodal task, therefore, we propose a high-performance, highly general Dual-Router Dynamic Framework (DRDF), which is highly modular and applicable in both multimodal and unimodal tasks. DRDF receives text and image information through text router and image router in Dual-Router, and determines the importance of modal information through MWF-Layer and fuses them into fused weights to guide subsequent experts to fuse to solve the problem. We performed DRDF extensions for multiple backbones on multimodal and unimodal datasets, and our DRDF outperforms all the baselines.

In future work we will continue to optimize the importance determination mechanism of DRDF and test DRDF in a wider range of domains.

\newpage

\bibliographystyle{ACM-Reference-Format}
\bibliography{sample-base}
\end{document}